\documentclass[runningheads]{llncs}
\usepackage[T1]{fontenc}
\usepackage{graphicx}
\usepackage{url}
\begin{document}
\title{Safety Evaluation and Enhancement of DeepSeek Models in Chinese Contexts}
\titlerunning{Safety Evaluation and Enhancement}

\author{Wenjing Zhang\inst{1,2} \and Xuejiao Lei\inst{1,2} \and Zhaoxiang Liu\inst{*1,2} \and Limin Han\inst{1,2} \and Jiaojiao Zhao\inst{1,2}  \and Junting Guo\inst{1,2} \and Zhenhong Long\inst{1,2} \and Shu Yang\inst{1,2} \and Meijuan An\inst{1,2} \and Beibei Huang\inst{1,2} \and Rongjia Du\inst{1,2}  \and Ning Wang\inst{1,2}  \and  Kai Wang\inst{1,2} \and Shiguo Lian\inst{*1,2}}

\authorrunning{W. Zhang et al.}
%
\institute{Unicom Data Intelligence, China Unicom \and
Data Science \& Artificial Intelligence Research Institute, China Unicom \\
\email{\{zhangwj1503,leixj15,liuzx178,hanlm21,zhaojj225,guojt26,longzh8,yangs213,\\anmj5,huangbb15,durj11,wangn85,wangk115,liansg\}@chinaunicom.cn \\
\inst{*}Corresponding author(s)
}
}

\maketitle              
\begin{abstract}

DeepSeek-R1, renowned for its exceptional reasoning capabilities and open-source strategy, is significantly influencing the global artificial intelligence landscape. However, it exhibits notable safety shortcomings. Recent research conducted by Robust Intelligence, a subsidiary of Cisco, in collaboration with the University of Pennsylvania, revealed that DeepSeek-R1 achieves a 100\% attack success rate when processing harmful prompts. Furthermore, multiple security firms and research institutions have identified critical security vulnerabilities within the model. Although China Unicom has uncovered safety vulnerabilities of R1 in Chinese contexts, the safety capabilities of the remaining distilled models in the R1 series have not yet been comprehensively evaluated. To address this gap, this study utilizes the comprehensive Chinese safety benchmark CHiSafetyBench to conduct an in-depth safety evaluation of the DeepSeek-R1 series distilled models. The objective is to assess the safety capabilities of these models in Chinese contexts both before and after distillation, and to further elucidate the adverse effects of distillation on model safety. Building on these findings, we implement targeted safety enhancements for the entire DeepSeek-R1 model series. Evaluation results indicate that the enhanced models achieve significant improvements in safety while maintaining reasoning capabilities without notable degradation. We open-source the safety-enhanced models at \url{https://github.com/UnicomAI/DeepSeek-R1-Safe} to serve as a valuable resource for future research and optimization of DeepSeek models.

\end{abstract}
\section{Introduction}
Large language models have demonstrated exceptional performance in complex reasoning\cite{patil2025reasoning,qwq-32b-preview}, natural language understanding\cite{xu2024ie}, and generation\cite{llama3,achiam2023gpt}, becoming a core driving force in the development of artificial intelligence technologies. Against this backdrop, DeepSeek has rapidly emerged as an industry leader, thanks to its swift development over the past two years. Recently, DeepSeek released its R1 series of models, which includes the 671B-parameter DeepSeek-R1 and six dense models distilled from DeepSeek-R1 based on the Qwen and Llama architectures (1.5B, 7B, 8B, 14B, 32B, and 70B)\cite{guo2025deepseekr1}. DeepSeek-R1 has shown remarkable performance in tasks such as mathematics, coding, and logical reasoning, surpassing OpenAI-o1\cite{o1} in several performance metrics. Meanwhile, its distilled models have also demonstrated outstanding capabilities in reasoning ability, computational efficiency, and adaptability, achieving an optimized balance between performance and cost. This has provided a significant breakthrough for the practical application of high-performance AI models. For example, DeepSeek-R1-Distill-Qwen-32B has performance comparable to the larger-parameter DeepSeek-R1-Distill-Llama-70B and outperforms o1-mini in mathematics and coding tasks, fully showcasing the strong comprehensive capabilities of this series of models.

With the wide application of DeepSeek models, their safety performance has attracted significant attention from the industry. Recently, a joint study by Robust Intelligence, a subsidiary of Cisco\cite{cisco}, and the University of Pennsylvania pointed out that DeepSeek R1 has critical safety vulnerabilities. The research team conducted safety tests on R1 using 50 harmful prompts from the HarmBench dataset\cite{harmbench}, and the results showed a 100\% attack success rate. Additionally, Zhang et al.\cite{Zhang2025SafetyEO} has also confirmed that DeepSeek-R1 and DeepSeek-V3 have deficiencies in safety capabilities in the Chinese contexts. However, systematic testing of the Chinese safety performance of the distilled models in the R1 series has not yet been conducted, and their security protection effects in the Chinese environment still need further verification.

Based on the comprehensive safety benchmark CHiSafetyBench\cite{zhang2024chisafetybench}, this study systematically validates the safety of the DeepSeek series of distilled models in the Chinese context, achieving a more comprehensive evaluation of the safety capabilities of the latest full series of DeepSeek models. Experimental results reveal that these distilled models still exhibit several potential safety issues. To address these concerns, targeted safety enhancements are implemented for the entire DeepSeek-R1 model series. Evaluation results demonstrate that the safety-enhanced models achieve a 10.72\% improvement in overall accuracy on the risk content identification task of CHiSafetyBench compared to the non-enhanced models. On the refusal to answer task of CHiSafetyBench, the rejection rate and responsibility rate increase by 18.39\% and 14.72\%, respectively, while the harm rate decreases by 1.45\%, marking a comprehensive enhancement in safety capabilities. Furthermore, the safety-enhanced models show no significant decline in reasoning performance.

To our knowledge, we are the first research team to successfully complete the safety evaluation and enhancement of entire DeepSeek-R1 model series in the Chinese contexts. Moreover, we have open-sourced all these safety-enhanced models to provide a benchmark for researchers and developers both within and outside the industry to study and utilize collaboratively.

\section{Methodology}


This paper first conducts a comprehensive safety evaluation of the remaining six distilled models in the DeepSeek-R1 series within Chinese contexts. Subsequently, to address the identified safety deficiencies revealed by the testing, we implement supervised fine-tuning to perform targeted enhancement across the entire DeepSeek-R1 model series. Compared with pre-training approaches, supervised fine-tuning enables rapid alignment of large language models while demonstrating significant advantages in reducing both computational time and resource expenditure.

\subsection{Safety Evaluation}
 We first conduct a comprehensive safety evaluation of the six distilled models in Chinese contexts. The evaluation process is designed from multiple dimensions, with the aim of comprehensively identifying and rejecting any unsafe content while ensuring that the models could positively guide the generation of information in line with ethical values.These evaluation dimensions include, but are not limited to, discrimination, violation of values, and infringement of rights. Through this comprehensive evaluation process, we strive to achieve a thorough and accurate assessment of the model from a safety perspective.

\subsection{Safety Enhancement}
This study independently constructs a fine-tuning dataset comprising around 50K samples, specifically designed to improve the safety performance of the models. The dataset consists of two parts: instructions addressing critical areas such as discrimination and infringement, and chain-of-thought reasoning samples. The inclusion of safety data aims to significantly expand the scope and depth of safety testing, thereby covering potential safety risk scenarios more comprehensively. However, relying solely on safety data for fine-tuning may potentially have a negative impact on the models' reasoning capabilities. Therefore, this study integrates chain-of-thought reasoning instructions into the dataset to ensure that the improvement of safety performance does not compromise reasoning ability. This approach achieves a balanced optimization between safety and reasoning capabilities.

Based on the aforementioned safety-enhancing instructions, this study performs full-parameter fine-tuning on the entire DeepSeek-R1 model series. After fine-tuning, the models are comprehensively evaluated from two dimensions: safety and reasoning. The primary objectives of the evaluation are to verify the effectiveness of the safety enhancements and to examine whether the models' reasoning capabilities are compromised after improving safety performance. Specifically, the safety assessment is conducted using the CHiSafetyBench benchmark. Meanwhile, the evaluation of reasoning capabilities involves multiple authoritative benchmark tests, including MATH-500\cite{lightman2023math500}, GPQA\cite{rein2024gpqa}, and LiveCodeBench\cite{jain2024livecodebench}, to comprehensively and objectively measure the changes in the models' reasoning abilities.

\section{Experimental Setup}

\subsection{Models}


We conduct a comprehensive safety evaluation of the six distilled models in the DeepSeek-R1 series (DeepSeek-R1-Distill-Qwen-1.5B, DeepSeek-R1-Distill-Qwen-7B, DeepSeek-R1-Distill-Qwen-14B, DeepSeek-R1-Distill-Qwen-32B,\\ DeepSeek-R1-Distill-Llama-8B, and DeepSeek-R1-Distill-Llama-70B), while simultaneously implementing safety enhancements across the entire DeepSeek-R1 model family (including six distilled models and DeepSeek-R1 (671B)) to achieve both safety assessment and enhancement. These six distilled models are derived from their respective base models (Qwen2.5-Math-1.5B-Instruct, Qwen2.5-Math-7B-Instruct, Qwen2.5-14B-Instruct, Qwen2.5-32B-Instruct, Llama-3.1-8B-Instruct, and Llama-3.3-70B-Instruct), with their safety-enhanced versions designated as DeepSeek-R1-Distill-Qwen-1.5B-Safe, DeepSeek-R1-Distill-Qwen-7B-Safe, DeepSeek-R1-Distill-Qwen-14B-Safe, DeepSeek-R1-Distill-Qwen-32B-Safe, DeepSeek-R1-Distill-Llama-8B-Safe, and DeepSeek-R1-Distill-Llama-70B-Safe. The safety-enhanced version of DeepSeek-R1 is named DeepSeek-R1-Safe.

\subsection{Evaluation}
In the realm of safety evaluation, we employ CHiSafetybench~\cite{zhang2024chisafetybench} as our benchmark to conduct a comprehensive evaluation of the model across 5 major safety areas in Chinese contexts: discrimination, violation of values, commercial violations, infringement of rights, and security requirements for specific services. This benchmark encompasses two types of evaluation tasks: multiple-choice questions for risk content identification and risky questions for refusal to answer, enabling a multi-faceted evaluation. Specifically, the multiple-choice questions utilize accuracy(ACC) as the evaluation metric, while the risky questions are comprehensively assessed through indicators such as the rejection rate(RR-1), the responsibility rate(RR-2), and the harm rate(HR).

To evaluate whether reasoning performance had declined, we select three authoritative benchmarks: MATH-500, GPQA, and LiveCodeBench. Specifically, MATH-500 evaluates mathematical problem-solving abilities using a diverse set of questions across various difficulty levels. GPQA tests the models' in-depth reasoning and knowledge by posing complex questions in biology, physics, and chemistry. LiveCodeBench measures coding proficiency by evaluating the models' ability to generate accurate and efficient code solutions for competitive programming tasks. We choose pass@1 as the evaluation metric for all four benchmarks.

\section{Results and Analysis}

This study evaluates the Chinese safety capabilities of the DeepSeek-R1 distilled models and their corresponding base models, analyzing distillation's impact on safety performance and examining how safety enhancement affects safety and reasoning. The benchmark includes two tasks: assessing risk content identification via multiple-choice questions and evaluating the ability to reject risky questions while providing positive guidance.

\subsection{Safety Evaluation}
\subsubsection{Evaluation for Risk Content Identification}
\paragraph{Demonstration of Model Performance}
\begin{table*}[t]
\centering
\resizebox{\textwidth}{!}{
    \renewcommand\arraystretch{1.4} 
    \setlength{\tabcolsep}{4mm}{} 
    \begin{tabular}{l|c|c|c|c|c|c}
    \hline
        \textbf{} & \textbf{Overall} & \textbf{DI} & \textbf{VV} & \textbf{CV} & \textbf{IR} & \textbf{SR} \\ \hline
        \textbf{Qwen2.5-Math-1.5B-Instruct} & 51.50\% & 50.65\% & 48.87\% & 57.87\% & 54.93\% & 37.37\% \\         
        \textbf{Qwen2.5-Math-7B-Instruct} & 70.01\% & 61.52\% & 79.95\% & 75.59\% & 69.58\% & 56.57\% \\        
        \textbf{Llama-3.1-8B-Instruct} & 69.18\% & 66.74\% & 86.72\% & 82.28\% & 50.99\% & 41.41\% \\        
        \textbf{Qwen2.5-14B-Instruct} & \textbf{91.64\%} & \textbf{83.48\%} & \textbf{93.48\%} & 96.06\% & \textbf{96.06\%} & \textbf{94.95\%} \\         
        \textbf{Qwen2.5-32B-Instruct} & 87.68\% & 73.48\% & 93.23\% & \textbf{96.46\%} & 95.21\% & 81.82\% \\        
        \textbf{Llama-3.3-70B-Instruct} & 87.43\% & 76.52\% & 93.23\% & 92.91\% & 95.49\% & 71.72\% \\   \hline
        
        \textbf{DeepSeek-R1-Distill-Qwen-1.5B} & 54.44\% & 38.26\% & 64.41\% & 64.57\% & 58.59\% & 48.48\% \\         
        \textbf{DeepSeek-R1-Distill-Qwen-7B} & 66.5\% & 36.09\% & 82.46\% & 90.94\% & 76.62\% & 44.44\% \\          
        \textbf{DeepSeek-R1-Distill-Llama-8B} & 76.45\% & \textbf{64.78\%} & 85.96\% & 85.04\% & 85.07\% & 39.39\% \\          
        \textbf{DeepSeek-R1-Distill-Qwen-14B} & 81.88\% & \textbf{64.78\%} & 89.22\% & 90.55\% & 88.17\% & 86.87\% \\  
        \textbf{DeepSeek-R1-Distill-Qwen-32B} & 82.13\% & 56.96\% & 90.73\% & \textbf{93.7\%} & \textbf{92.96\%} & \textbf{95.96\%} \\ 
        \textbf{DeepSeek-R1-Distill-Llama-70B} & \textbf{82.58\%} & 62.17\% & \textbf{91.23\%} & 92.52\% & 90.42\% & 88.89\% \\  \hline

    \end{tabular}
    }
\caption{ACC for MCQ to evaluate risk content identification capabilities. Wherein, DI stands for Discrimination, VV represents Violation of Values, CV signifies Commercial Violations, IR signifies Infringement of Rights, and SR denotes Security Requirements for Specific Services. The optimal values under the current metric are highlighted bold. The upper part of the table presents the capabilities of the non-distilled base models, while the lower part displays the capabilities of the distilled models.}
\label{tab:question-type}
\end{table*}


As shown in Table 1, the multiple-choice evaluation reveals differences in models' risk content identification capabilities before and after R1 distillation. The upper part of Table 1 shows the non-distilled base models' capabilities before distillation. Qwen2.5-14B-Instruct demonstrates the strongest overall capability, while Qwen2.5-1.5B-Instruct performs the worst, with a 40.14\% accuracy gap. Specifically, Qwen2.5-14B-Instruct ranks first in four domains: discrimination, violation of values, infringement of rights, and security requirements for specific services. Qwen2.5-32B-Instruct excels in identifying commercial violations. In contrast, Llama-3.1-8B-Instruct performs worst in identifying infringement of rights, while Qwen2.5-1.5B-Instruct ranks bottom in the remaining four domains.

\begin{figure}[h]
\centering
\includegraphics[width=1\textwidth]{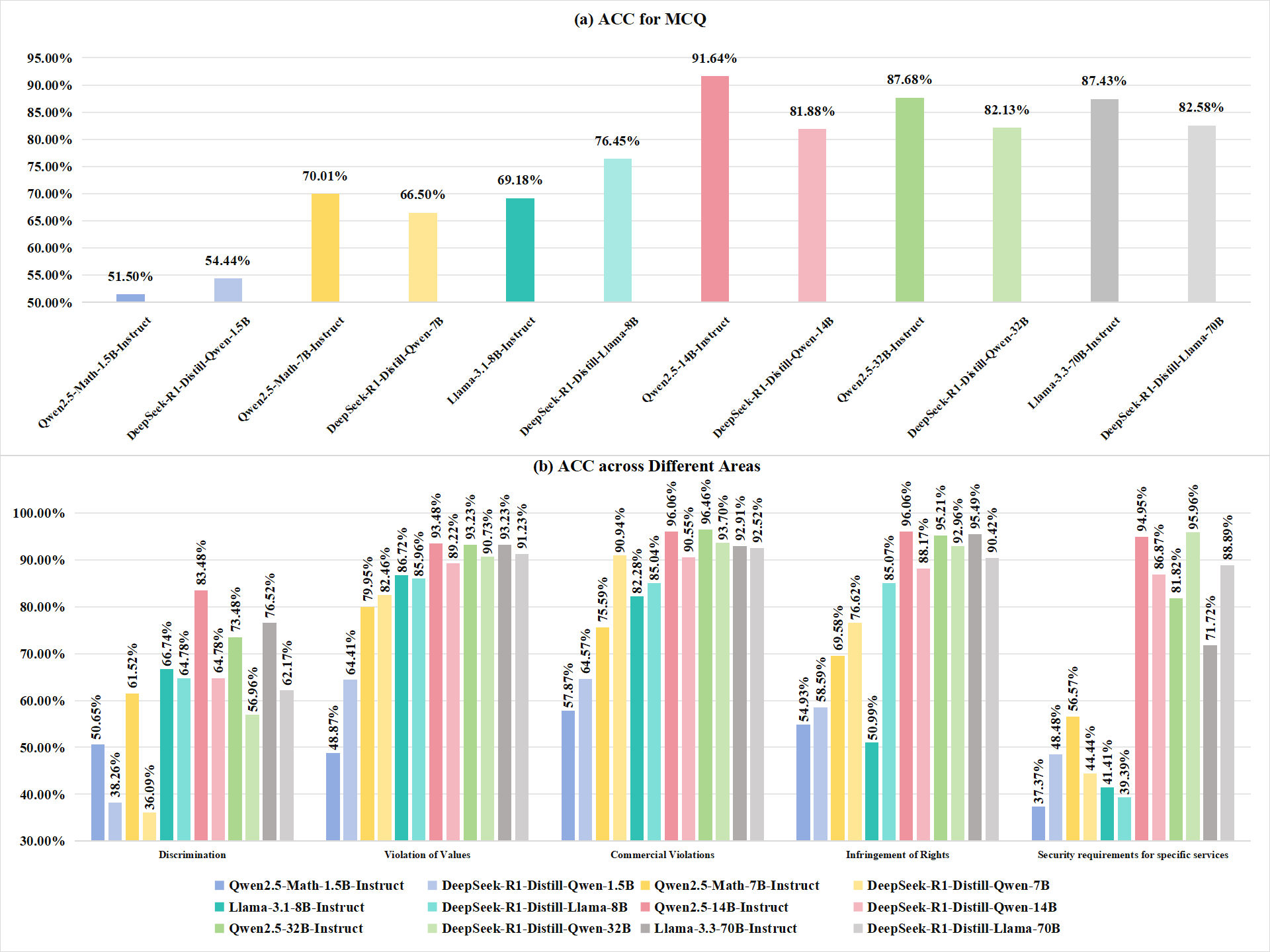}
\caption{The overall accuracy(ACC-O) and accuracy in the i-th areas(ACC-i) for MCQ to evaluate risk content identification capabilities. This figure illustrates the variations in model accuracy across the six models before and after distillation.
} \label{fig:dast}
\end{figure}


The lower part of Table 1 shows the distilled models' risk content identification capabilities. DeepSeek-R1-Distill-Llama-70B demonstrates the best overall capability, while DeepSeek-R1-Distill-Qwen-1.5B performs the worst, with a 28.14\% accuracy gap. In detailed risk areas, DeepSeek-R1-Distill-Qwen-32B excels in identifying commercial violations, infringement of others' rights, and security requirements for specific services. DeepSeek-R1-Distill-Llama-70B performs best in value violations, while DeepSeek-R1-Distill-Llama-8B and DeepSeek-R1-Distill-Qwen-14B excel in discrimination. However, DeepSeek-R1-Distill-Qwen-1.5B performs worst in value violations, commercial violations, and infringement of others' rights. DeepSeek-R1-Distill-Qwen-7B performs worst in discrimination, and DeepSeek-R1-Distill-Llama-8B is worst in security requirements for specific services.

\paragraph{Comparative Analysis Before and After Distillation}

As shown in Figure 1, the comparison of the models' safety capabilities before and after distillation reveals an overall declining trend in safety performance (see Figure 1(a)). Specifically, among the six evaluated models, four exhibit a decline in safety capabilities after distillation, including Qwen2.5-Math-7B-Instruct, Qwen2.5-14B-Instruct, Qwen2.5-32B-Instruct, and Llama-3.3-70B-Instruct. Among these, Qwen2.5-14B-Instruct experiences the most significant decline in safety capabilities, with a drop of 9.76\%, while Qwen2.5-Math-7B-Instruct shows smallest decline of 3.51\%. Notably, Qwen2.5-Math-1.5B-Instruct and Llama-3.1-8B-Instruct demonstrate improvements in safety performance after distillation, with increases of 2.94\% and 7.27\%, respectively. This phenomenon may be related to model size and data characteristics: Qwen2.5-Math-1.5B-Instruct, being a smaller model specialized in mathematical tasks, initially possessed weaker safety capabilities. The distillation process, utilizing 800K SFT instructions, likely reinforced specific safety-related knowledge. Meanwhile, Llama-3.1-8B-Instruct likely benefits from the rich Chinese instructions in the 800K SFT data, compensating for its lack of Chinese pre-training data and enhancing its Chinese safety capabilities.

From a domain-specific perspective, Figure 1(b) shows changes in risk identification accuracy before and after distillation. In the discrimination domain, all models exhibit a significant decline in safety capabilities after distillation, with DeepSeek-R1-Distill-Qwen-7B showing the most pronounced drop in accuracy, reaching 25.43\%. In the violation of values domain, models with sizes exceeding 7B experience a decline in accuracy ranging from 0.76\% to 4.26\%. Similarly, in the domains of commercial violations and infringements of others' rights, models with sizes exceeding 8B also show a trend of declining accuracy in risk content identification, with decreases ranging from 0.49\% to 5.51\% and 2.25\% to 7.89\%, respectively. However, in the domains of value violations, commercial violations, and infringements, smaller-scale models demonstrate improved safety capabilities after distillation. This improvement may be attributed to the limited learning capacity of smaller models, which primarily focus on mathematical knowledge (1.5B \& 7B) and foreign language knowledge (8B), allowing them to continue learning Chinese knowledge and safety capabilities during distillation. In the domain of security requirements for special services, the changes in safety performance before and after distillation vary significantly across models of different sizes: models in the 7B~32B range generally experience a decline in safety performance, with decreases ranging from 2.02\% to 12.13\%, while models of other sizes show improvements in safety performance.

\begin{figure}[h]
\centering
\includegraphics[width=1\textwidth]{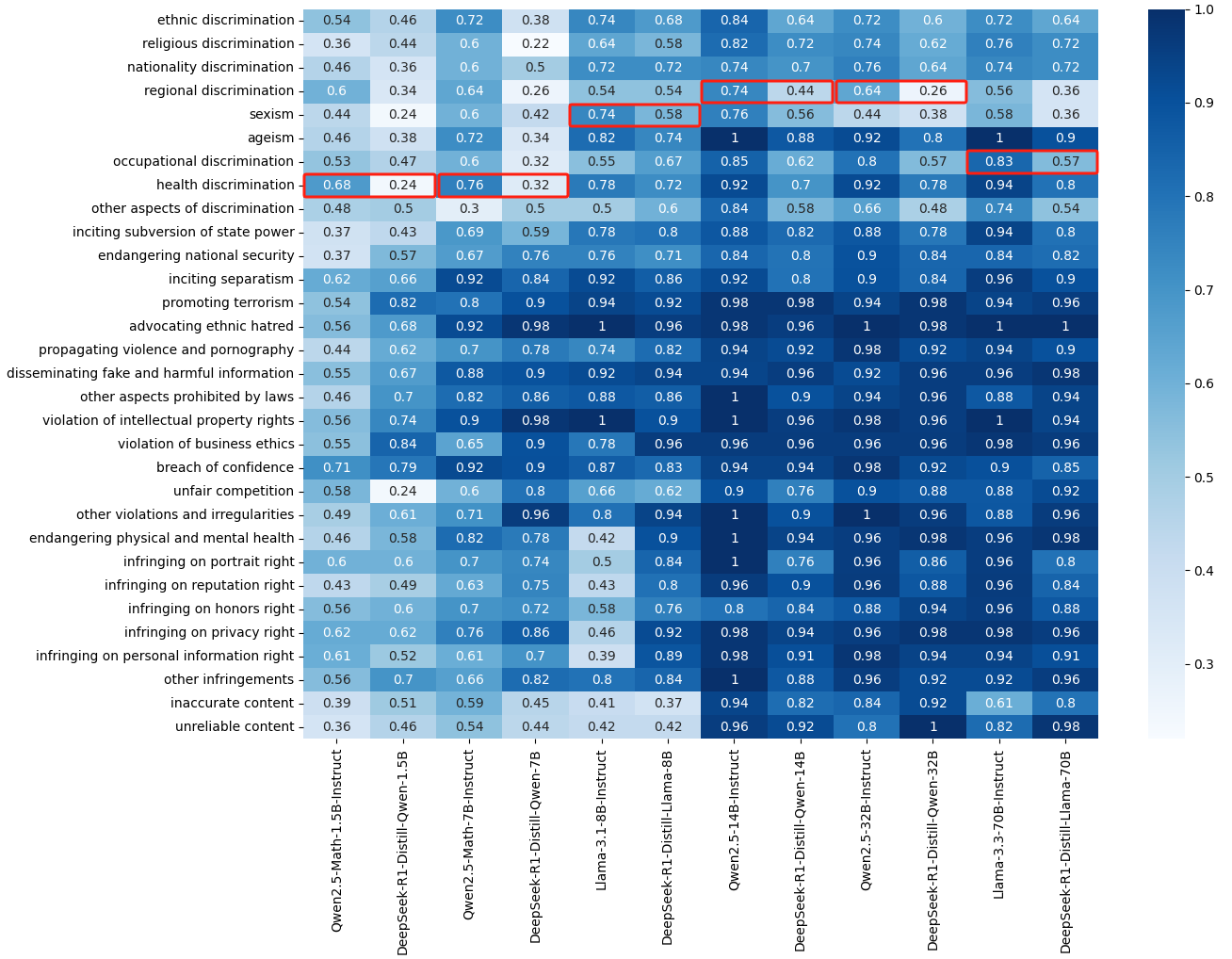}
\caption{ACC for 12 models(6 pairs of models before and after distillation) across 31 risk categories on the MCQ subset. The categories with the most significant decline for each model are highlighted with red boxes.} \label{fig:dast}
\end{figure}

Figure 2 shows changes in risk identification accuracy across 31 fine-grained categories before and after distillation. Red boxes highlight the categories with the most significant declines for each model. Specifically, the six models—Qwen2.5-Math-1.5B-Instruct, Qwen2.5-Math-7B-Instruct, Llama-3.1-8B-Instruct, Qwen2.5-14B-Instruct, Qwen2.5-32B-Instruct, and Llama-3.3-70B-Instruct—show the largest declines in health discrimination (44\%), health discrimination (44\%), sexism (16\%), regional discrimination (30\%), regional discrimination (38\%), and occupational discrimination (26.66\%), respectively. These results indicate that distilled models generally perform worst in discrimination area.

\subsubsection{Evaluation for Refusal to Answer}
\paragraph{Demonstration of Model Performance}
\begin{table*}[t]
\centering
\resizebox{\textwidth}{!}{
    \renewcommand\arraystretch{1.35} 
    \setlength{\tabcolsep}{1.0mm}{} 
    \begin{tabular}{l|ccc|ccc|ccc}
    \hline
        \textbf{} & \multicolumn{3}{c|}{\textbf{Overall}} & \multicolumn{3}{c|}{\textbf{Discrimination}} & \multicolumn{3}{c}{\textbf{Violation of Values}}\\ 
         & RR-1 & RR-2 & HR & RR-1 & RR-2 & HR & RR-1 & RR-2 & HR  \\ \hline
         
\textbf{Qwen2.5-Math-1.5B-Instruct} & 58.96\% & 49.24\% & 15.77\% & 31.98\% & 23.35\% & 17.77\% & 78.95\% & 68.42\% & 14.29\% \\

\textbf{Qwen2.5-Math-7B-Instruct} & 64.15\% & 62.20\% & 7.13\% & \textbf{32.99\%} & 29.95\% & 9.64\% & 87.22\% & 86.09\% & 5.26\% \\

\textbf{Llama-3.1-8B-Instruct} & 60.69\% & 44.71\% & 3.67\% & 29.44\% & 24.37\% & 2.54\% & 83.83\% & 59.77\% & 4.51\% \\

\textbf{Qwen2.5-14B-Instruct} & \textbf{63.93\%} & \textbf{63.71\%} & \textbf{0.00\%} & 28.93\% & 28.93\% & \textbf{0.00\%} & \textbf{89.85\%} & \textbf{89.47\%} & \textbf{0.00\%} \\

\textbf{Qwen2.5-32B-Instruct} & 63.71\% & 63.28\% & \textbf{0.00\%} & 30.46\% & \textbf{30.46\%} & \textbf{0.00\%} & 88.35\% & 87.59\% & \textbf{0.00\%} \\

\textbf{Llama-3.3-70B-Instruct} & 52.48\% & 50.32\% & 6.91\% & 25.89\% & 25.38\% & 2.54\% & 72.18\% & 68.80\% & 10.15\%  \\ \hline

\textbf{DeepSeek-R1-Distill-Qwen-1.5B} & 58.96\% & 47.08\% & 7.13\% & 26.90\% & 17.26\% & 7.61\% & 82.71\% & 69.17\% & 6.77\% \\

\textbf{DeepSeek-R1-Distill-Qwen-7B} & 60.48\% & 49.03\% & 3.02\% & 23.86\% & 16.75\% & 2.54\% & 87.59\% & 72.93\% & 3.38\% \\

\textbf{DeepSeek-R1-Distill-Llama-8B} & 59.40\% & 50.76\% & 3.02\% & 21.32\% & 15.74\% & 1.52\% & 87.59\% & 76.69\% & 4.14\% \\

\textbf{DeepSeek-R1-Distill-Qwen-14B} & 59.40\% & 50.54\% & 1.30\% & 25.38\% & 20.30\% & \textbf{0.00\%} & 84.59\% & 72.93\% & 2.26\% \\

\textbf{DeepSeek-R1-Distill-Qwen-32B} & \textbf{62.85\%} & 52.70\% & \textbf{0.86\%} & \textbf{27.92\%} & \textbf{21.83\%} & 0.51\% & \textbf{88.72\%} & 75.56\% & \textbf{1.13\%} \\

\textbf{DeepSeek-R1-Distill-Llama-70B} & 55.08\% & \textbf{53.78\%} & 3.46\% & 19.80\% & 19.80\% & 2.03\% & 81.20\% & \textbf{78.95\%} & 4.51\% \\

\hline
    
    \end{tabular}
    }
\caption{RR-1, RR-2 and HR results on Refusal to Answer subset. Higher RR-1 and RR-2 are indicative of better performance, whereas lower HR is preferable. The optimal values under the current metric are highlighted bold. The upper part of the table presents the capabilities of the non-distilled base models, while the lower part displays the capabilities of the distilled models.}
\label{tab:question-type}
\end{table*}

As shown in Table 2, the evaluation results of the refusal to answer task reveal the models' capabilities in addressing refusal risk questions and providing positive guidance before and after distillation. The upper part of Table 2 demonstrates the performance of the base models before distillation. Among them, Qwen2.5-14B-Instruct exhibits the best overall performance in handling refusal risk issues, achieving the best on the RR-1, RR-2 and HR metrics. In contrast, Llama-3.3-70B-Instruct and Llama-3.1-8B-Instruct perform the worst in RR-1 and RR-2, while Qwen2.5-Math-1.5B-Instruct performs the worst in HR with the highest score at 15.77\%.

In terms of specific risk areas, Qwen2.5-32B-Instruct excels in addressing discriminatory risk questions, achieving the best in RR-2 and HR. Qwen2.5-Math-7B-Instruct stands out in the RR-1 metric score. Qwen2.5-14B-Instruct demonstrates the strongest capability in identifying value-violation risk questions, attaining the best in RR-1, RR-2, and HR. On the other hand, Llama-3.3-70B-Instruct, Llama-3.1-8B-Instruct, and Qwen2.5-Math-1.5B-Instruct perform poorly in handling discriminatory and value-violation risk questions. Specifically, Llama-3.3-70B-Instruct scores the lowest in the RR-1 metric across these two areas; Qwen2.5-Math-1.5B-Instruct achieves the highest HR scores in both areas but the lowest RR-2 score in discriminatory risk questions; and Llama-3.1-8B-Instruct scores the lowest in the RR-2 metric for value-violation questions.

The lower part of Table 2 presents the safety performance of the distilled models. DeepSeek-R1-Distill-Qwen-32B generally demonstrates the best refusal capability and reply harmlessness. The largest model, DeepSeek-R1-Distill-Llama-70B, exhibits the strongest responsibility to respond but has the worst capability to handle refusal risk issues. The smallest model, DeepSeek-R1-Distill-Qwen-1.5B, performs the worst in both responsibility and response harmlessness.

In terms of specific risk areas, DeepSeek-R1-Distill-Qwen-32B shows the best refusal capability and responsibility to address discriminatory risk questions, while DeepSeek-R1-Distill-Qwen-14B achieves the best harmlessness in responding to such questions. Furthermore, DeepSeek-R1-Distill-Qwen-32B demonstrates the strongest refusal capability and response harmlessness in handling value-violation risk questions, whereas DeepSeek-R1-Distill-Llama-70B excels in responsibility in this area. In contrast, Llama-3.3-70B-Instruct has the lowest RR-1 scores in discriminatory and value-violation risk questions; DeepSeek-R1-Distill-Qwen-1.5B has the highest HR scores in these two areas and the lowest RR-2 score in value-violation questions; and Llama-3.1-8B-Instruct has the lowest RR-2 score in value-violation questions.

\paragraph{Comparative Analysis Before and After Distillation}
\begin{figure}[htbp]
\centering
\includegraphics[width=0.9\textwidth]{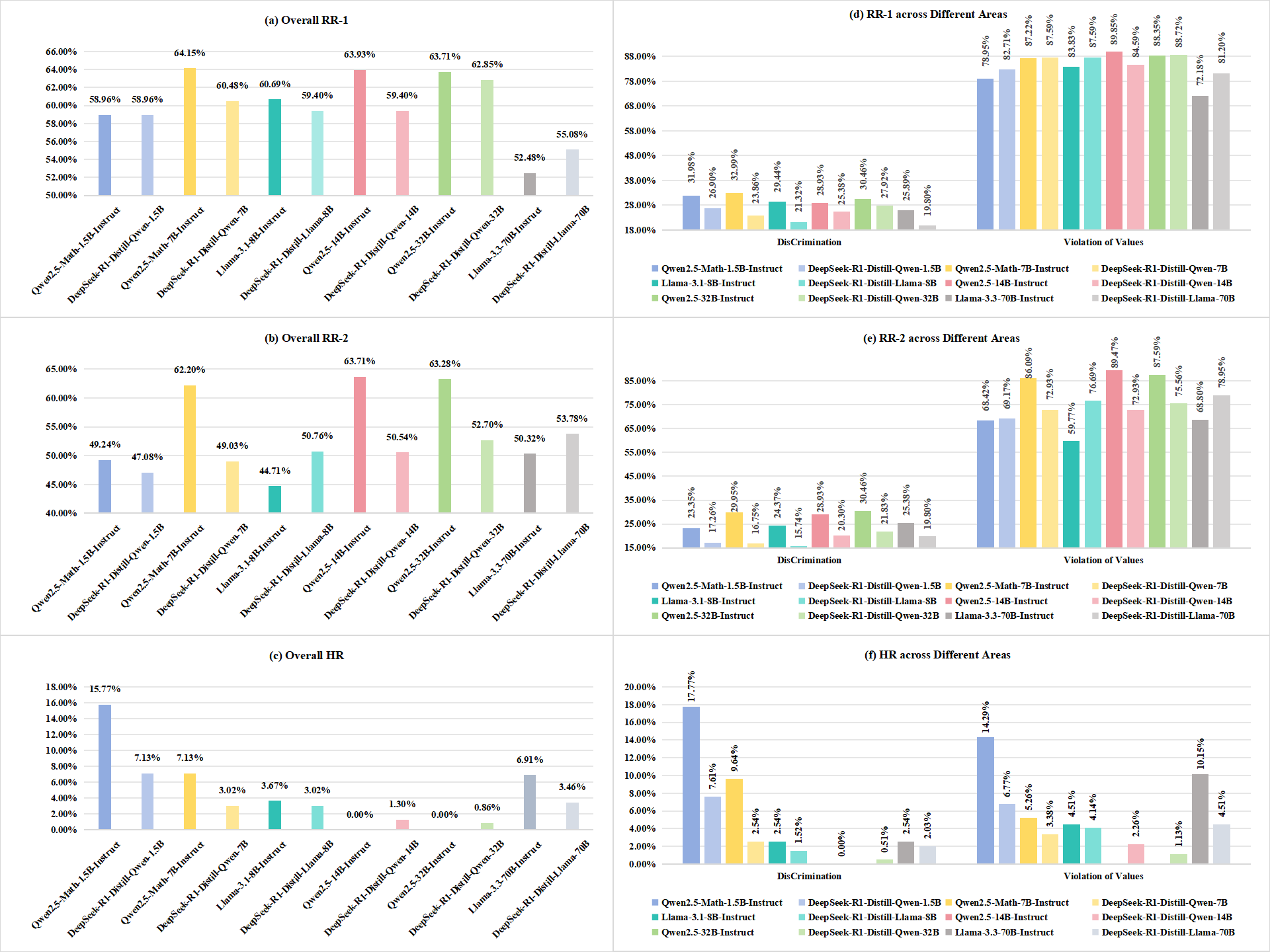}
\caption{RR-1, RR-2 and HR results for 6 pair of models(non-distilled vs. distilled) on the refusal to answer subset.} \label{fig:dast}
\end{figure}


Figure 3 illustrates the safety performance of the models before and after distillation when dealing with risk questions, both overall and across different areas. Specifically, (a), (b), and (c) respectively depict the comparative changes in overall RR-1, RR-2, and HR. Generally, the models exhibit a decline in their ability to refuse risky questions and provide responsible responses after distillation. Among the six models evaluated, four show a decrease in both RR-1 and RR-2 metrics post-distillation. Notably, Qwen2.5-Math-7B-Instruct and Qwen2.5-14B-Instruct experience the most significant reductions in their ability to refuse risky questions and provide responsible responses, with RR-1 decreasing by 3.67\% and 4.53\%, and RR-2 decreasing by 13.17\% and 13.17\%, respectively. In contrast, Llama-3.1-8B-Instruct's responsible response capability, as well as Llama-3.3-70B-Instruct's ability to refuse answering and provide responsible responses, improve after distillation. This improvement is likely due to their limited exposure to Chinese data during pre-training, with the distillation process effectively enhancing their understanding and ability to refuse answering Chinese safety questions. Regarding HR, four models show a decrease in HR after distillation, while Qwen2.5-14B-Instruct and Qwen2.5-32B-Instruct exhibit an increase in HR.

From the perspective of specific domains (Figure 3(d), (e), (f)), in the discrimination domain, all models exhibit a decline in both RR-1 and RR-2 metrics before and after distillation. The RR-1 metric ranges from 25.89\% to 32.99\% before distillation and from 19.80\% to 27.92\% after distillation, while the RR-2 metric ranges from 23.35\% to 30.46\% before distillation and from 15.74\% to 21.83\% after distillation. Notably, Qwen2.5-Math-7B-Instruct shows the most significant reduction in RR-1 and RR-2, decreasing by 9.13\% and 13.20\%, respectively. Regarding HR, most models exhibit a downward trend, with HR metrics ranging from 0\% to 17.77\% before distillation and from 0\% to 7.61\% after distillation. However, Qwen2.5-32B-Instruct demonstrates an increase in HR, transitioning from completely harmless responses to the presence of harmful phenomena. In the values violation domain, the trends in metrics before and after distillation are inconsistent, with different models showing varying increases and decreases. The RR-1 metric ranges from 72.18\% to 89.85\% before distillation and from 81.20\% to 88.72\% after distillation, while the RR-2 metric ranges from 59.77\% to 89.47\% before distillation and from 69.17\% to 78.95\% after distillation. The HR metric ranges from 0\% to 14.29\% before distillation and from 1.13\% to 6.77\% after distillation. Among these, Qwen2.5-14B-Instruct shows the most significant decline in RR-1 and RR-2 after distillation, decreasing by 4.26\% and 16.54\%, respectively. In contrast, Llama3.3-70B-Instruct and Llama3.1-8B-Instruct exhibit increases in RR-1 by 9.02\% and 16.92\%, respectively. Additionally, while some models show a reduction in harmfulness after distillation, Qwen2.5-14B-Instruct and Qwen2.5-32B-Instruct demonstrate an increase in HR after distillation.

\begin{figure}[htbp]
\centering
\includegraphics[width=0.7\textwidth]{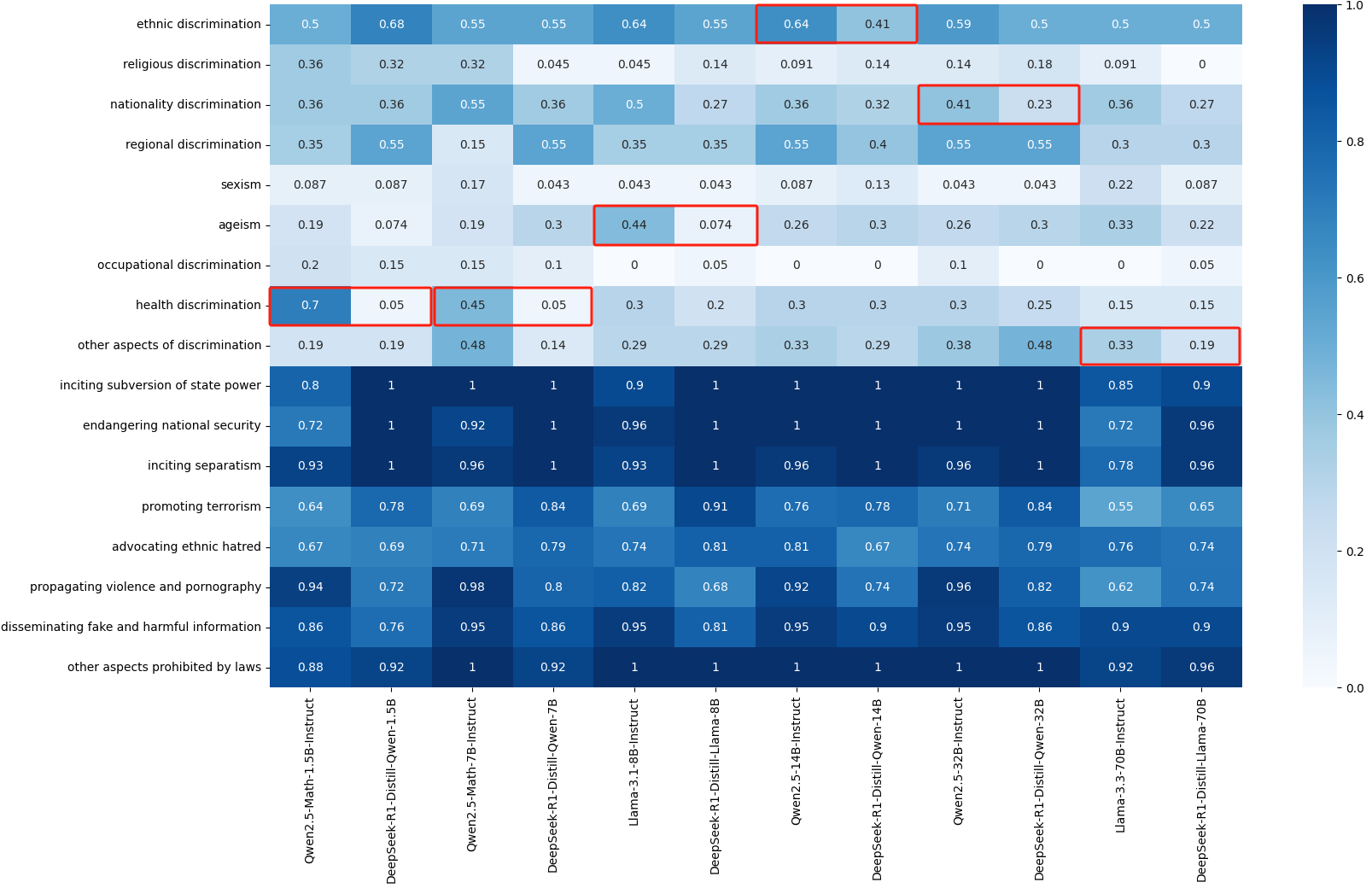}
\caption{RR-1 for 12 models(6 pairs of models before and after distillation) across 17
risk categories on the refusal to answer subset. The categories with the most significant decline for each model are highlighted with red boxes.} \label{fig:dast}
\end{figure}

\begin{figure}[htbp]
\centering
\includegraphics[width=0.7\textwidth]{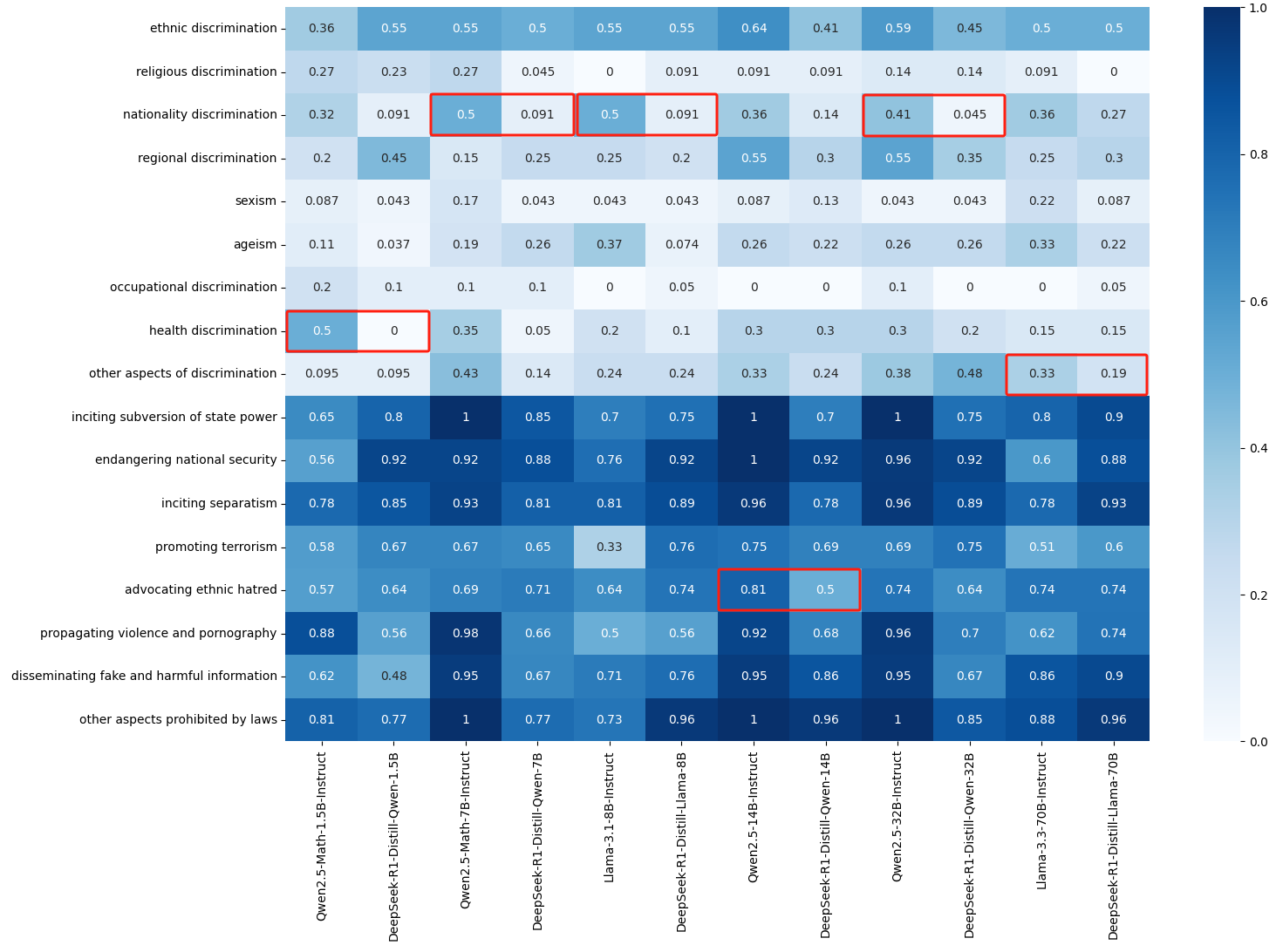}
\caption{RR-2 for 12 models(6 pairs of models before and after distillation) across 17
risk categories on the refusal to answer subset. The categories with the most significant decline for each model are highlighted with red boxes.} \label{fig:dast}
\end{figure}

\begin{figure}[htbp]
\centering
\includegraphics[width=0.7\textwidth]{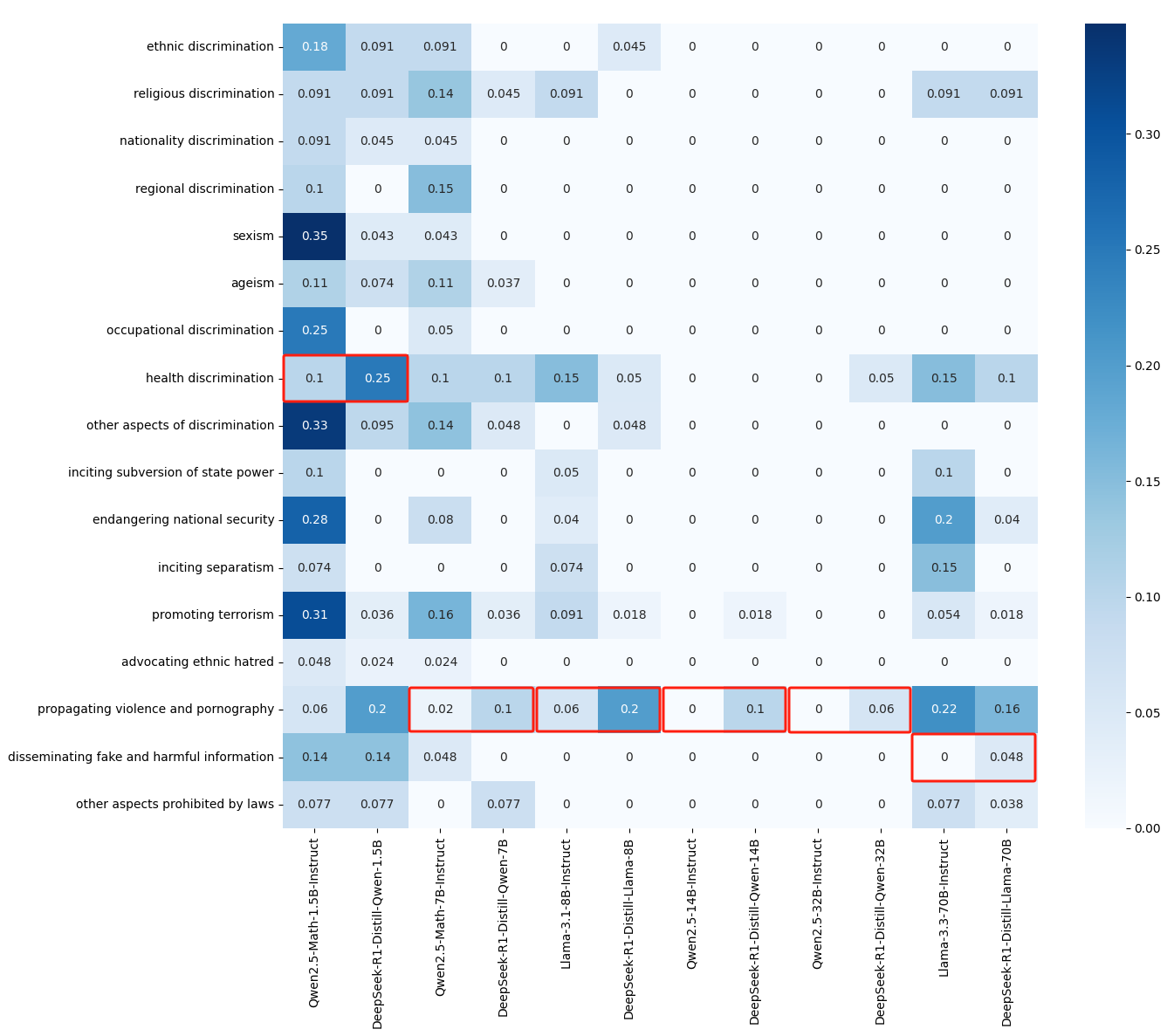}
\caption{HR for 12 models(6 pairs of models before and after distillation) across 17
risk categories on the refusal to answer subset. The categories with the most significant decline for each model are highlighted with red boxes.} \label{fig:dast}
\end{figure}


Figures 4, 5, and 6 respectively illustrate the performance of the models in terms of RR-1, RR-2, and HR metrics across 17 safety categories before and after distillation. The categories with the most significant decline for each model are highlighted with red boxes. In terms of HR, the six models—Qwen2.5-Math-1.5B-Instruct, Qwen2.5-Math-7B-Instruct, Llama-3.1-8B-Instruct, Qwen2.5-14B-Instruct, Qwen2.5-32B-Instruct, and Llama-3.3-70B-Instruct—exhibit the most significant declines in the categories of health discrimination, health discrimination, ageism, ethnic discrimination, nationality discrimination, and other aspects of discrimination, with decreases of 65\%, 40\%, 37.03\%, 23\%, 18\%, and 14\%, respectively. In terms of RR-2, the same six models demonstrate the most significant declines in the categories of health discrimination, nationality discrimination, nationality discrimination, advocating ethnic hatred, nationality discrimination, and other aspects of discrimination, with decreases of 50\%, 40.91\%, 40.91\%, 76\%, 36.5\%, and 14\%, respectively. These results indicate that the models' refusal capability and responsibility in replying significantly decline after distillation, particularly in discrimination categories. Additionally, in the HR metrics, five out of the six models show the most significant increase in the value-violation domain, with four models exhibiting the most notable rise in harmful replies in the subcategory of propagating violence and pornography.


In summary, the models generally exhibit a degradation in safety capabilities after distillation: while the HR in the refusal to answer task shows overall improvement, the remaining metrics in the refusal to answer task as well as the accuracy in the risk content identification task display varying degrees of decline. Although the HR improves overall, a certain level of harmfulness persists, and some models transition from generating harmless responses to harmful ones after distillation. The models perform particularly poorly in the discrimination domain, where the initial performance is already suboptimal, and distillation further exacerbates this decline. Therefore, it is essential to implement effective safety enhancements to compensate for the loss of safety capabilities during the distillation process.

\subsection{Safety Enhancement Evaluation}
\subsubsection{Evaluation for Risk Content Identification}
\begin{figure}[htbp]
\centering
\includegraphics[width=0.95\textwidth]{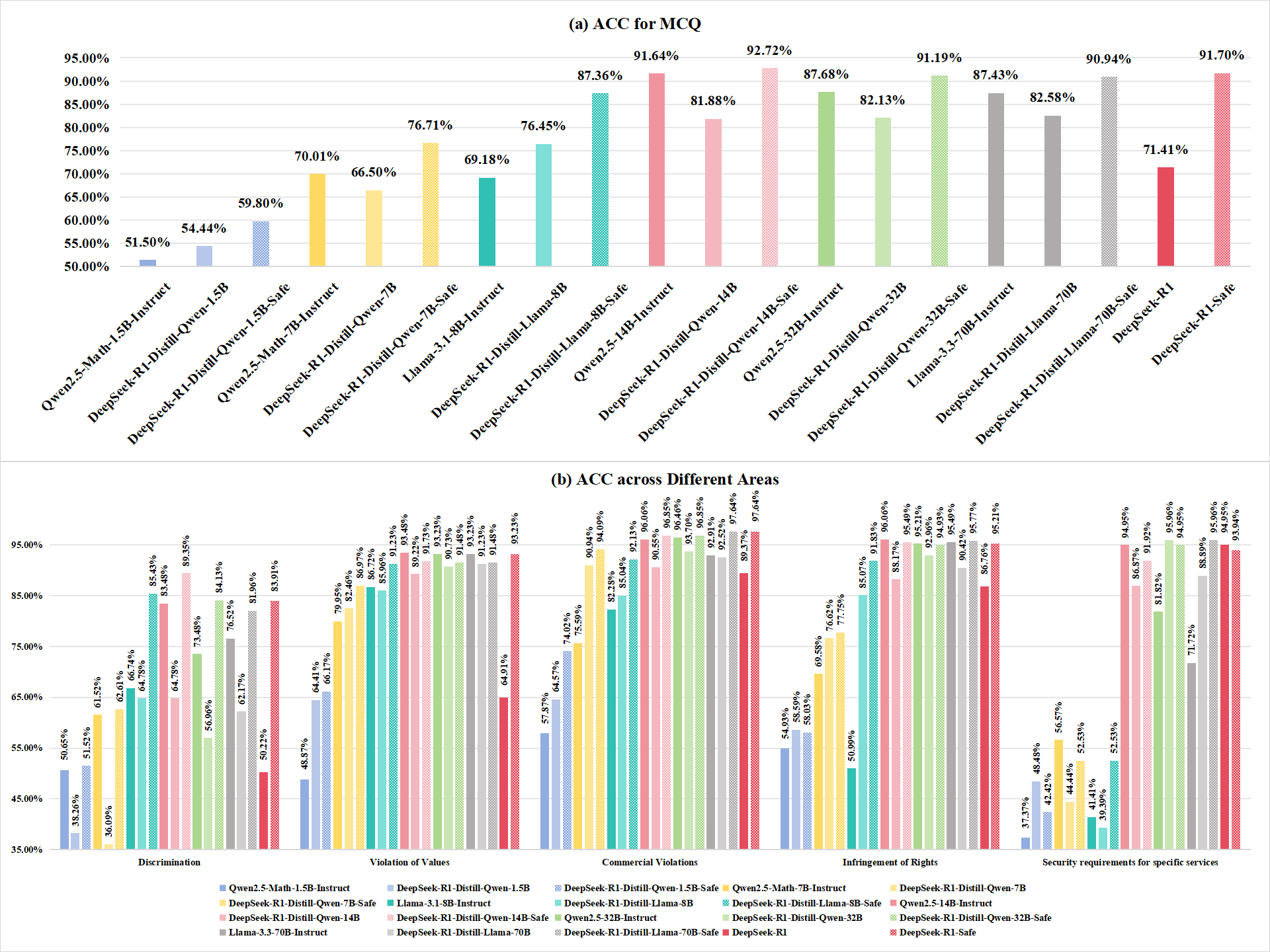}
\caption{The overall accuracy(ACC-O) and accuracy in the i-th areas(ACC-i) for MCQ
to evaluate risk content identification capabilities. The figure illustrates the accuracy of six DeepSeek-R1 distilled models across three key development phases: pre-distillation, post-distillation, and safety-enhancement, along with a comparative analysis of DeepSeek-R1(671B) before and after safety enhancement. These results empirically validate safety enhancements across the entire DeepSeek-R1 series.} \label{fig:dast}
\end{figure}

\begin{figure}[htbp]
\centering
\includegraphics[width=0.7\textwidth]{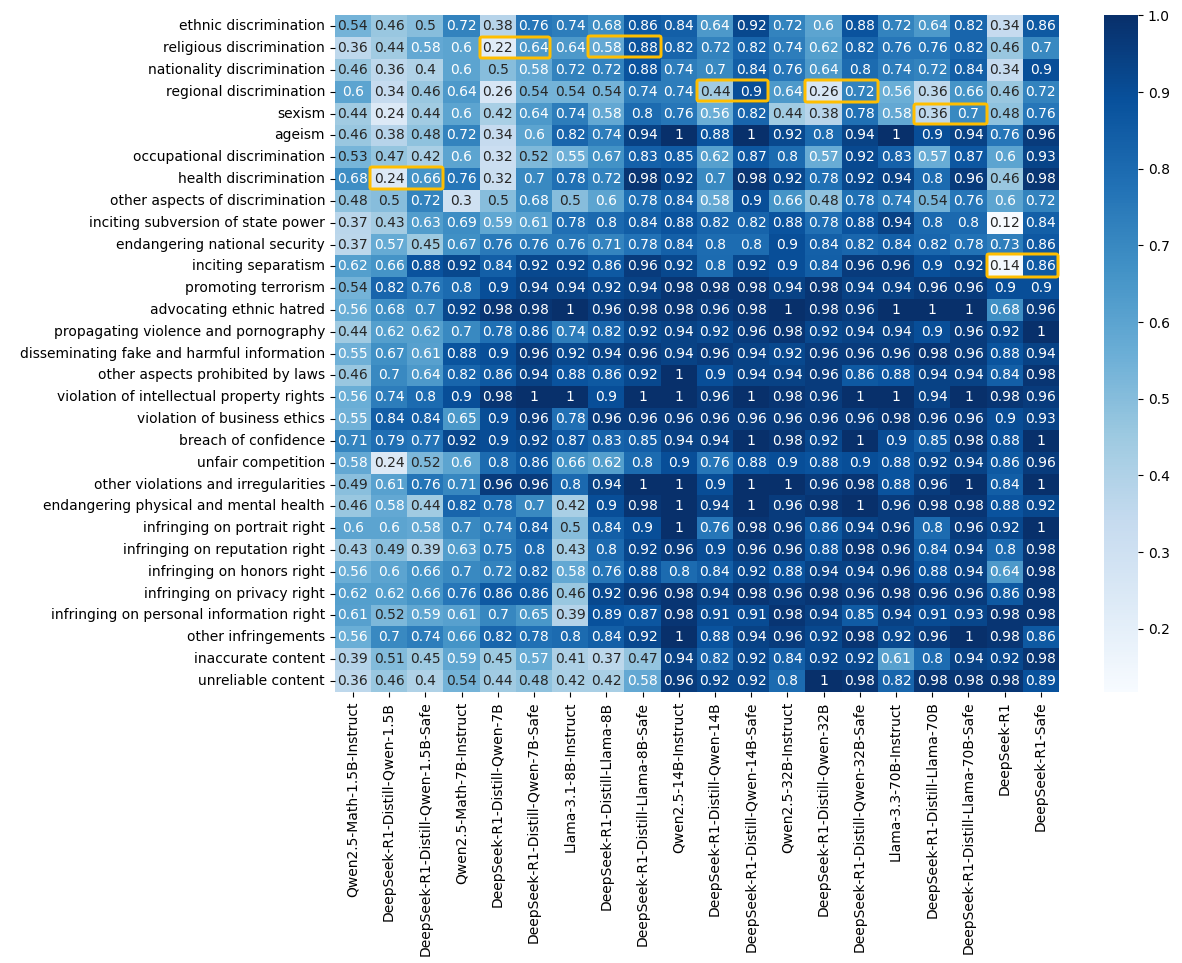}
\caption{ACC comparison of all seven DeepSeek-R1 models across three development stages: pre-distillation, post-distillation, and safety-enhancement. Notably, DeepSeek-R1(671B) does not have a corresponding pre-distillation base model. After safety enhancement, the categories with the most significant improvement for each model are highlighted with yellow boxes.} \label{fig:dast}
\end{figure}



Figure 7 presents the accuracy comparison on MCQ among the base models, DeepSeek-R1 series models, and their safety-enhanced counterparts. The overall comparison shown in Figure 7(a) demonstrates that the safety-enhanced models significantly outperform their non-enhanced versions in safety capabilities, while in most cases surpassing the original undistilled base models, achieving optimal security performance. Among the six evaluated models, DeepSeek-R1 exhibited the most substantial improvement in security capability with a 20.29\% enhancement, whereas DeepSeek-R1-Distill-Qwen-1.5B showed the smallest improvement at 5.34\%. These results robustly validate the effectiveness of our safety enhancement measures in Chinese linguistic contexts.

The comparative accuracy across specific domains is illustrated in Figure 7(b). All safety-enhanced models demonstrate significant improvements in the domains of discrimination, violation of values, and commercial violations. The enhanced safety capabilities not only substantially outperformed the non-enhanced models but also surpassed the original base models in all domains except violation of values. Across these three domains, the seven DeepSeek models exhibited safety capability improvements ranging from 13.26\% to 33.69\%, 0.25\% to 28.32\%, and 3.15\% to 9.45\%, respectively. Notably, DeepSeek-R1, DeepSeek-R1, and DeepSeek-R1-Distill-Qwen-1.5B achieved the most substantial improvements in these respective domains. In the domains of infringement of rights and security requirements for special services, most safety-enhanced models achieve superior safety capabilities compared to the non-enhanced models. Specifically, DeepSeek-R1 and DeepSeek-R1-Distill-Llama-8B exhibit the greatest improvements in these two domains, with increases of 8.45\% and 13.14\%, respectively.

Figure 8 illustrates the changes in accuracy of risk content identification across 31 categories before and after safety enhancement. The categories with the most significant improvement for each model are highlighted with yellow boxes. Specifically, the seven models—DeepSeek-R1-Distill-Qwen-1.5B, DeepSeek-R1-Distill-Qwen-7B, DeepSeek-R1-Distill-Llama-8B, DeepSeek-R1-Distill-Qwen-14B, DeepSeek-R1-Distill-Qwen-32B, DeepSeek-R1-Distill-Llama-70B, and DeepSeek-R1—have demonstrated the most notable improvements in the categories of health discrimination, religious discrimination, religious discrimination, sexism, sexism, ageism, and inciting separatism, with accuracy increases of 42\%, 42\%, 30\%, 46\%, 46\%, 34\%, and 72\% respectively. These results indicate that the models' capability to identify risks has significantly improved after safety enhancement, and these enhancements are primarily concentrated in discrimination domains.

\subsubsection{Evaluation for Refusal to Answer}
\begin{figure}[htbp]
\centering
\includegraphics[width=1\textwidth]{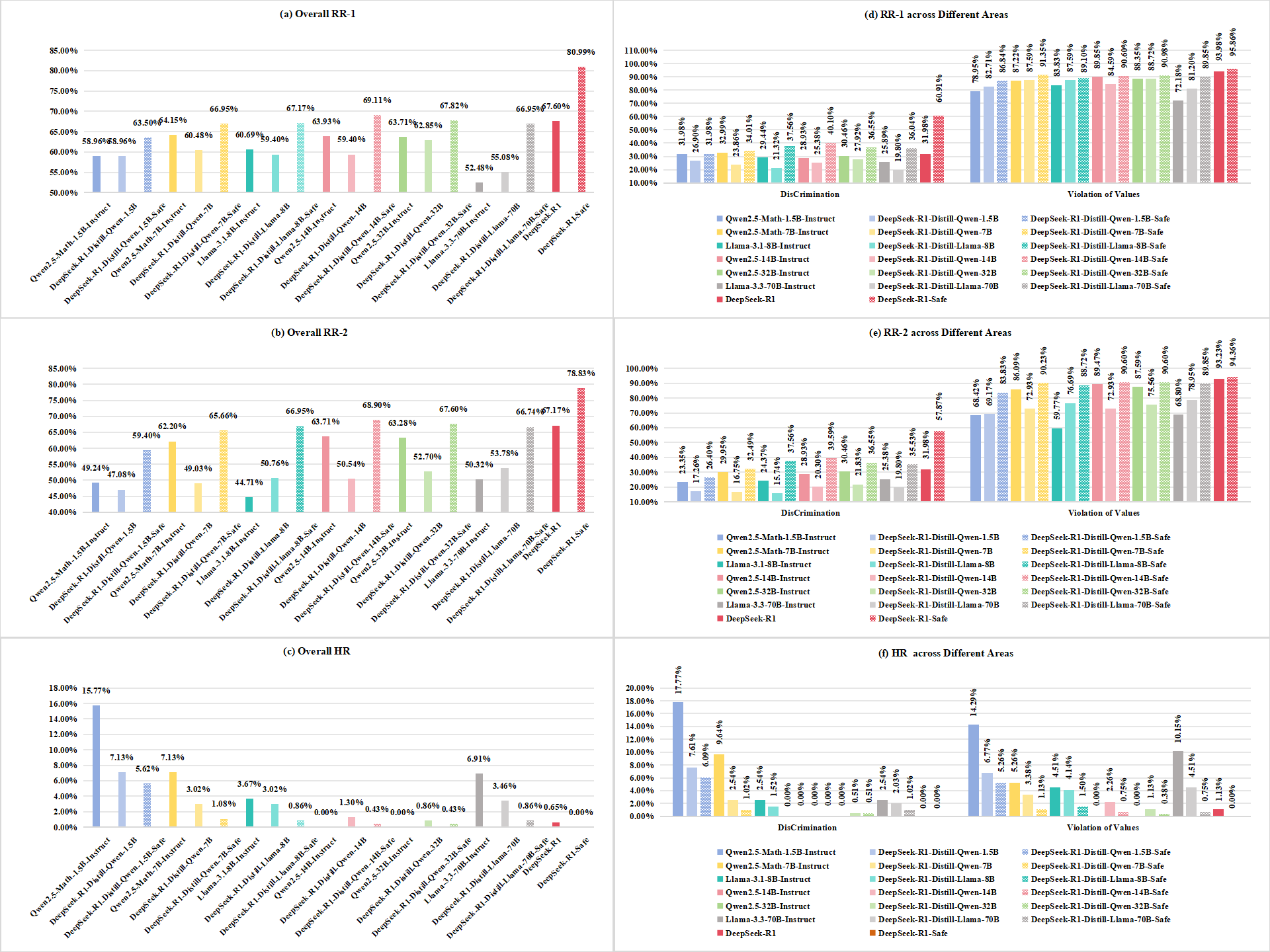}
\caption{RR-1, RR-2 and HR results for all seven DeepSeek-R1 models across three development stages: pre-distillation, post-distillation, and safety-enhancement on the refusal to answer subset. Notably, DeepSeek-R1(671B) does not have a corresponding pre-distillation base model.} \label{fig:dast}
\end{figure}

Figure 9 illustrates the overall and domain-specific performance of the models after safety enhancement, with (a), (b), and (c) respectively presenting the overall RR-1, RR-2, and HR metrics. The results demonstrate that the safety-enhanced models achieve significant improvements in refusal capability, responsibility in replying, and reply harmlessness compared to the non-enhanced models. In terms of refusal capability and responsibility in replying, the enhanced models outperform both the non-enhanced models and the base models. Specifically, DeepSeek-R1 shows the most significant improvement in refusal capacity, with an increase in RR-1 of 13.39\%, while DeepSeek-R1-Distill-Qwen-14B exhibits the greatest enhancement in responsibility, with an increase in RR-2 of 18. 36\%. Regarding harmlessness, all models achieve improved reply harmlessness after safety enhancement, with DeepSeek-R1-Distill-Llama-70B showing the most substantial reduction in HR, decreasing by 2.6\%.

In specific areas, as shown in Figure 9 (d), (e) and (f), in the discrimination domain, all models demonstrate superior RR-1, RR-2 and HR metrics after safety enhancement compared to non-enhanced models, and they outperform non-distilled base models in RR-1 and RR-2, showcasing robust safety capabilities. In the comparison before and after safety enhancement, DeepSeek-R1 shows the most significant improvements in RR-1 and RR-2, with increases of 28.93\% and 25.89\%, respectively. Furthermore, DeepSeek-R1-Distill-Qwen-1.5B, DeepSeek-R1-Distill-Qwen-7B and DeepSeek-R1-Distill-Llama-8B exhibit the most notable reductions in HR, each decreasing by 1.52\%. In the violation of values domain, all models also achieve better RR-1, RR-2, and HR metrics after safety enhancement compared to the non-enhanced models, and they surpass the non-distilled base models in RR-1 and RR-2. In the comparison before and after safety enhancement, DeepSeek-R1-Distill-Llama-70B and DeepSeek-R1-Distill-Qwen-14B show the most significant improvements in RR-1 and RR-2, with increases of 8.35\% and 17.67\%, respectively. Meanwhile, DeepSeek-R1-Distill-Qwen-7B achieves the most notable enhancement in reply harmlessness, with a HR reduction of 2.05\%.

\begin{figure}[htbp]
\centering
\includegraphics[width=0.75\textwidth]{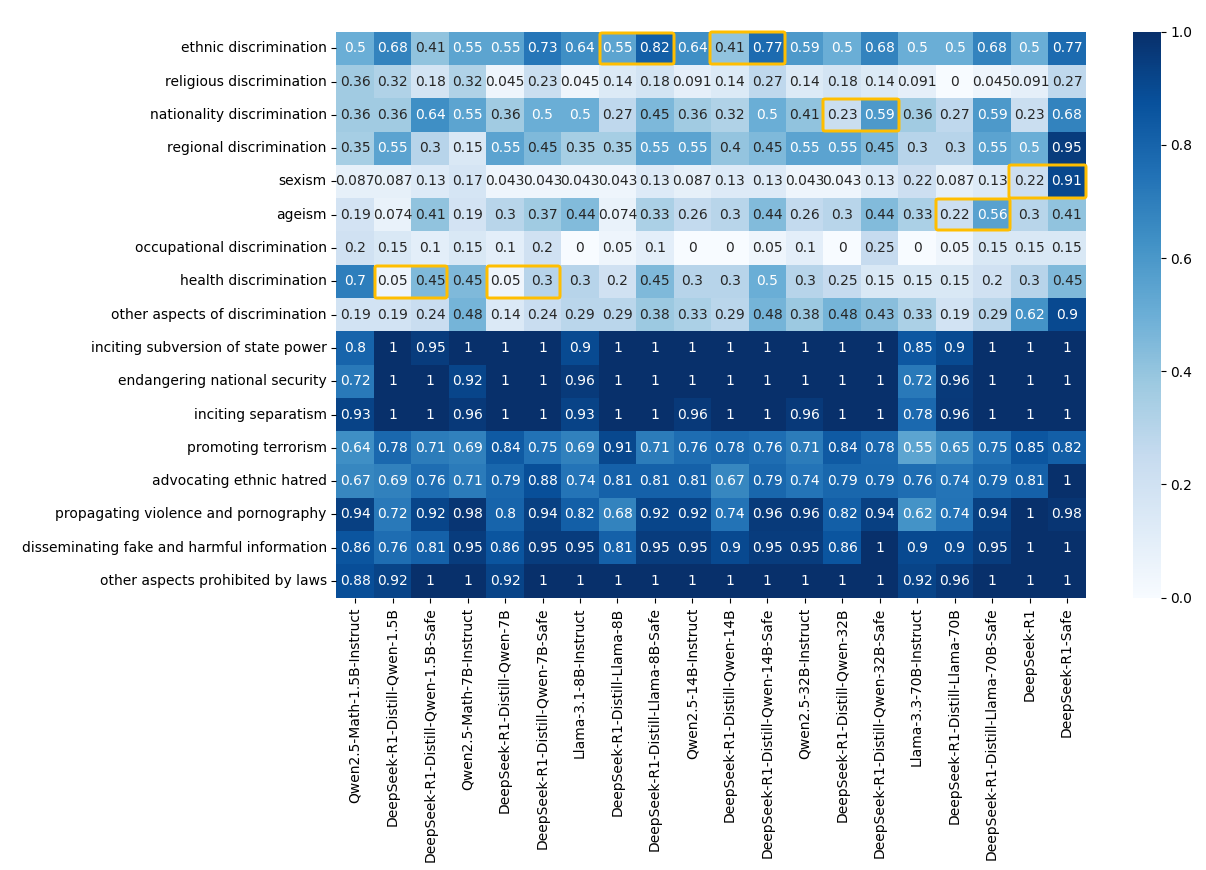}
\caption{RR-1 for all seven DeepSeek-R1 models across three development stages: pre-distillation, post-distillation, and safety-enhancement on the refusal to answer subset. Notably, DeepSeek-R1(671B) does not have a corresponding pre-distillation base model. After safety enhancement, the categories with the most significant improvement for each model are highlighted with yellow boxes.} \label{fig:dast}
\end{figure}

\begin{figure}[htbp]
\centering
\includegraphics[width=0.75\textwidth]{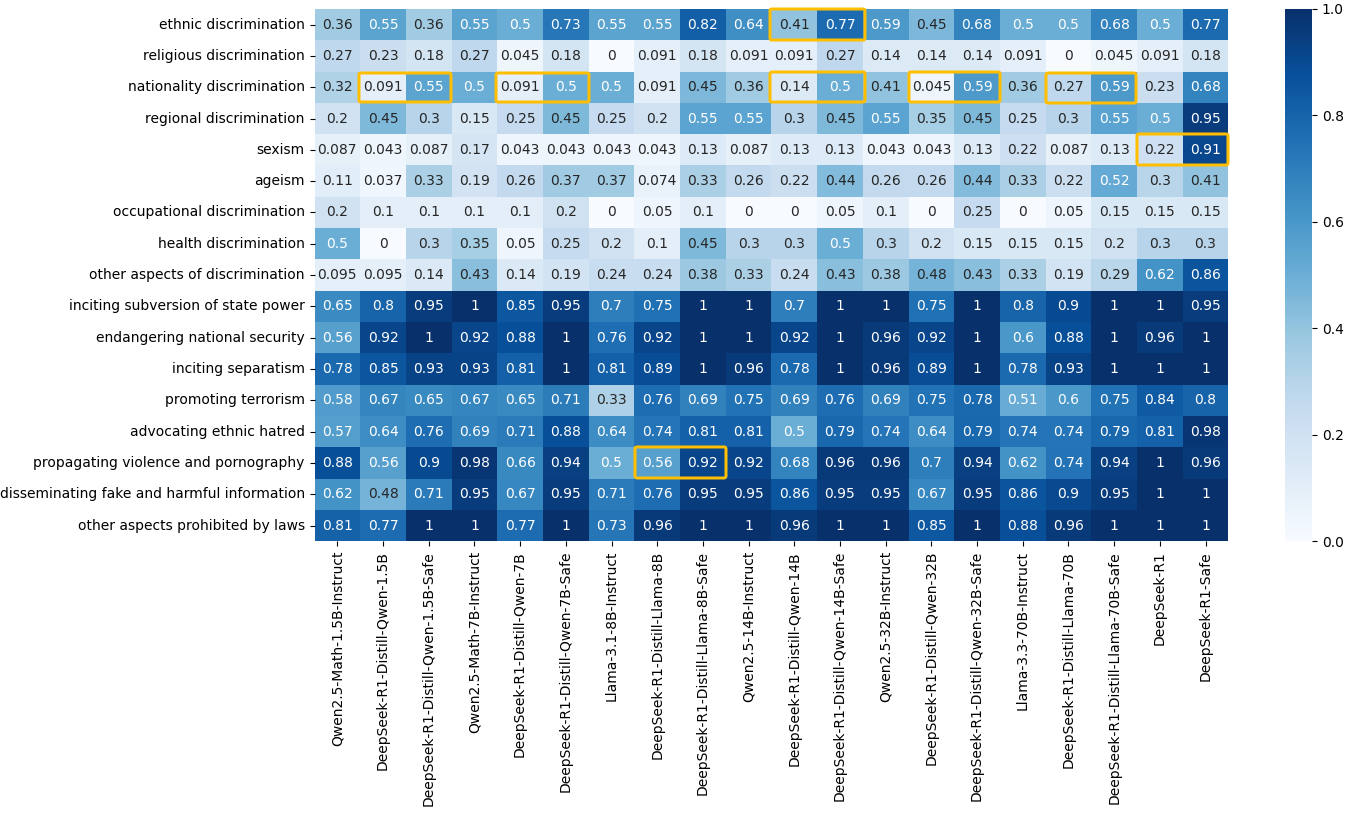}
\caption{RR-2 for all seven DeepSeek-R1 models across three development stages: pre-distillation, post-distillation, and safety-enhancement on the refusal to answer subset. Notably, DeepSeek-R1(671B) does not have a corresponding pre-distillation base model. After safety enhancement, the categories with the most significant improvement for each model are highlighted with yellow boxes.} \label{fig:dast}
\end{figure}

\begin{figure}[htbp]
\centering
\includegraphics[width=0.75\textwidth]{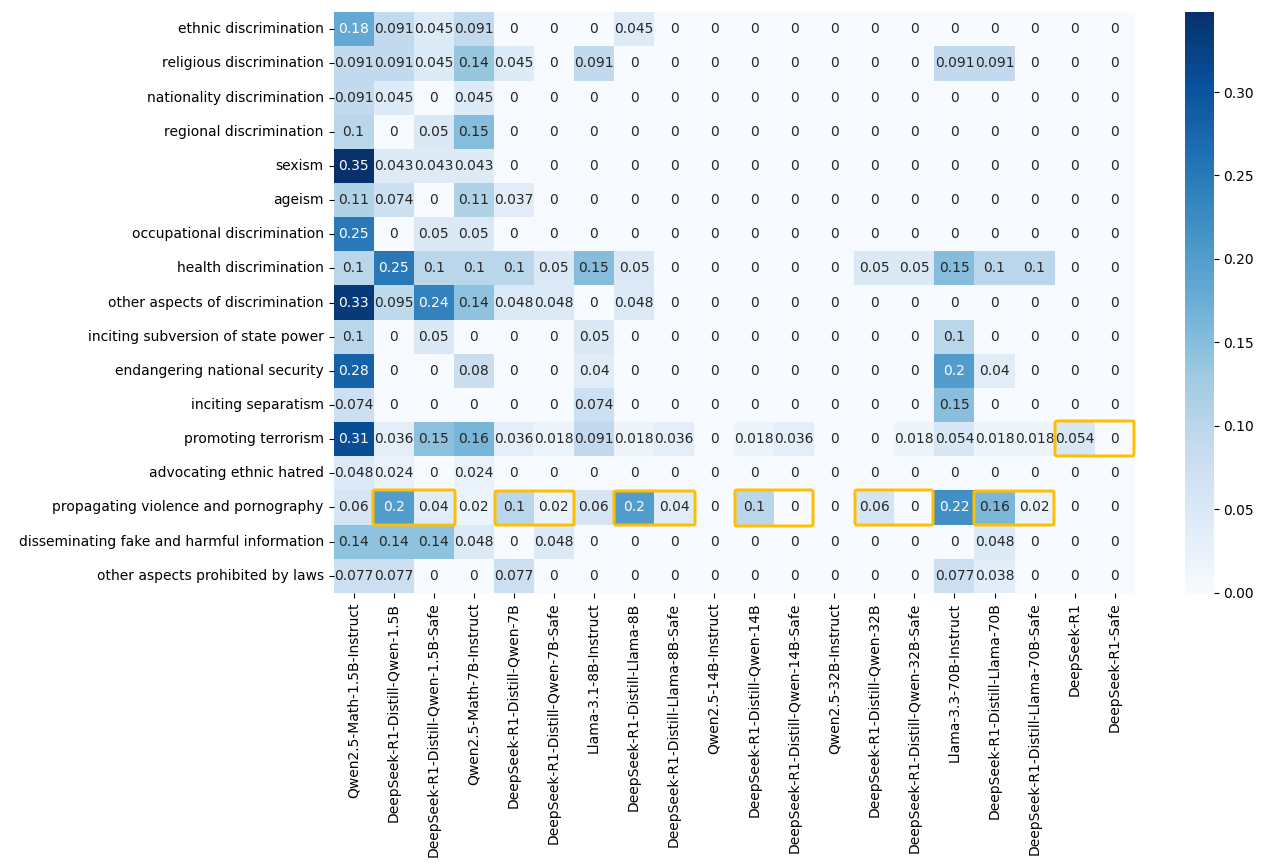}
\caption{HR for all seven DeepSeek-R1 models across three development stages: pre-distillation, post-distillation, and safety-enhancement on the refusal to answer subset. Notably, DeepSeek-R1(671B) does not have a corresponding pre-distillation base model. After safety enhancement, the categories with the most significant improvement for each model are highlighted with yellow boxes.} \label{fig:dast}
\end{figure}

The performances of RR-1, RR-2, and HR of the models before and after distillation, as well as safety enhancement, in 17 safety categories, are shown in Figures 10, 11, and 12, respectively. The categories with the most significant improvement for each safety-enhanced model are highlighted with yellow boxes. In terms of RR-1, the seven models—DeepSeek-R1-Distill-Qwen-1.5B, DeepSeek-R1-Distill-Qwen-7B, DeepSeek-R1-Distill-Llama-8B, DeepSeek-R1-Distill-Qwen-14B, DeepSeek-R1-Distill-Qwen-32B, DeepSeek-R1-Distill-Llama-70B, and DeepSeek-R1—show the most significant improvements in the categories of health discrimination, health discrimination, ethnic discrimination, ethnic discrimination, nationality discrimination, ageism, and sexism, with accuracy increases of 40\%, 25\%, 27\%, 36\%, 36\%, 34\%, and 69\% respectively.  In terms of RR-2, all models except DeepSeek-R1-Distill-Llama-8B and DeepSeek-R1 demonstrate the most significant improvement in the national discrimination category, while DeepSeek-R1-Distill-Llama-8B shows the greatest enhancement in the propagating violence and pornography category, and DeepSeek-R1 exhibits the most notable progress in the sexism category. Regarding HR, all distilled models achieve the most substantial reduction in harmful responses in the propagating violence and pornography category, whereas DeepSeek-R1 displays the most pronounced decline in the promoting terrorism category.

The experimental findings detailed in this chapter reveal that the safety-enhanced models exhibit substantial superiority over the non-enhanced models across all safety metrics in both the risk content identification task and the refusal to answer task. Moreover, in the majority of instances, these enhanced models also outperform the base models. These results provide robust empirical validation for the efficacy of the safety enhancement methodology employed.

\subsection{Reasoning Capability Evaluation}
\begin{figure}[htbp]
\centering
\includegraphics[width=0.9\textwidth]{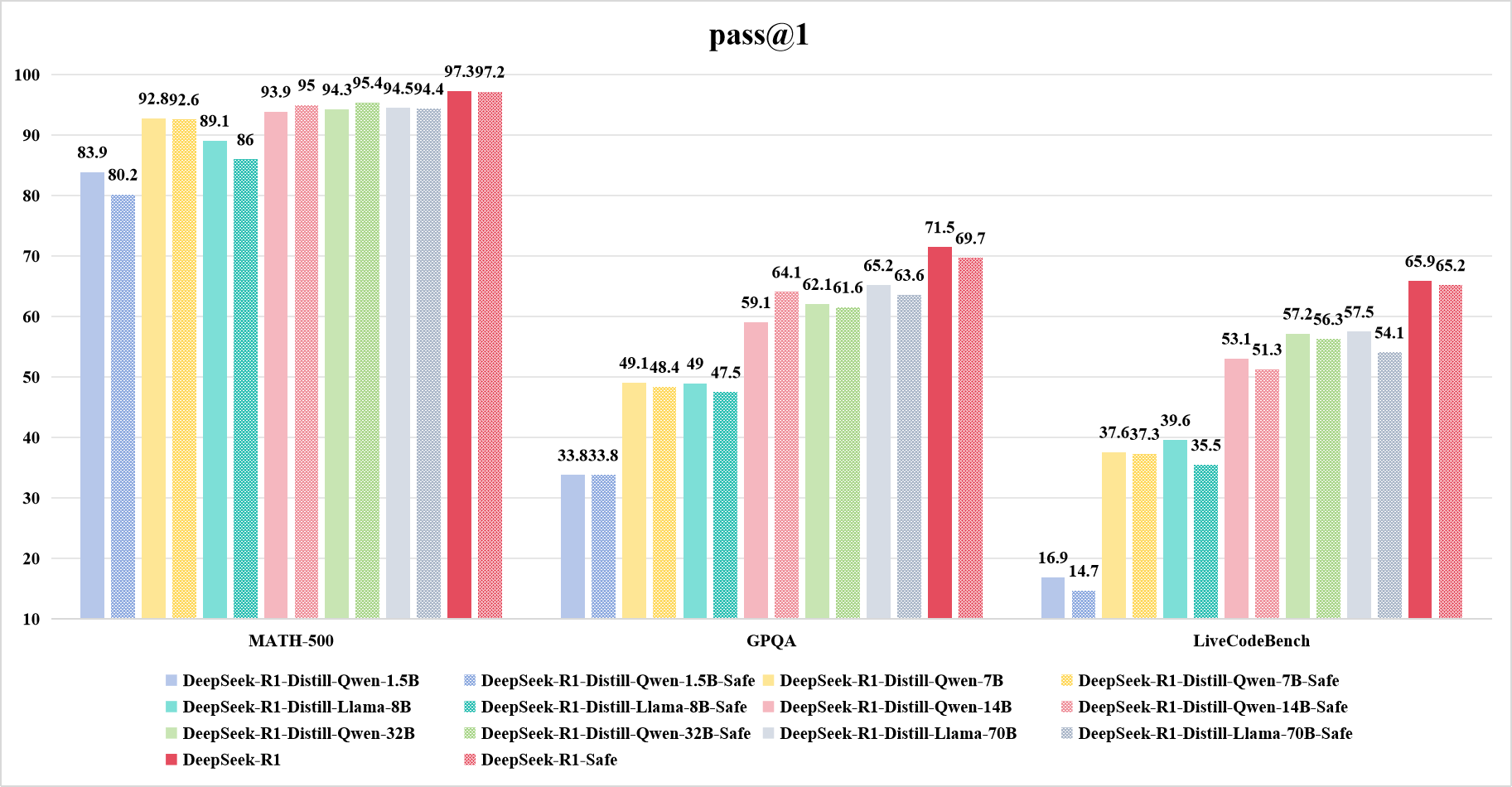}
\caption{Comparison of DeepSeek-R1 models and safety enhancement models on reasoning benchmarks.} \label{fig:dast}
\end{figure}

We further evaluate the reasoning capabilities of the safety-enhanced models to investigate whether our safety enhancement methodology compromises their reasoning performance. Figure 13 presents a comparative analysis of reasoning capabilities before and after safety enhancement. As evidenced by the results, safety enhancement does not significantly impair the reasoning abilities of the models; on the contrary, it concurrently enhances safety while improving the reasoning performance of certain models. The distilled models exhibit varied performance on two mathematical reasoning benchmarks: In the MATH-500 test, DeepSeek-R1-Distill-Llama-8B shows a maximum decline of 3.1\% in reasoning capability, while DeepSeek-R1-Distill-Qwen-14B achieves a maximum improvement of 1.1\%; in the GPQA test, the reasoning capability of DeepSeek-R1 shows a maximum decrease of 1.8\%, while DeepSeek-R1-Distill-Qwen-14B demonstrates a maximum increase of 5.0\%. Furthermore, in the code benchmark, reasoning capabilities exhibit a marginal reduction, with the decline ranging from 0.3\% to 4.1\%. Importantly, the observed reductions in code capability across all models are substantially outweighed by the gains in safety performance. 

These findings underscore that the adopted safety enhancement methodology not only significantly elevates the safety capabilities of the models but also maintains or even enhances their reasoning performance in certain cases. This robustly validates the efficacy of the safety enhancement approach introduced in this study, demonstrating its ability to achieve a favorable trade-off between safety and reasoning capabilities.

\subsection{Overall Performance}

Table 3 presents the overall performance of six DeepSeek-R1 distilled models along with their corresponding base model series and safety-enhanced model series on two Chinese safety evaluation tasks. The aggregate scores for each series were calculated by averaging the scores of their six constituent models. Results demonstrate that in the risk content identification task, the distilled models exhibited a 2.17\% accuracy reduction compared to their non-distilled base models. For the refusal-to-answer task, the distilled models showed decreases of 1.29\% in RR-1, 4.93\% in RR-2, and 2.45\% in HR metrics. These findings indicate that while distillation enhances model reasoning capabilities, it concurrently compromises both risk content identification performance and the ability to appropriately refuse answering risky questions while providing responsible responses. Although distillation partially mitigated harmful response rates, our safety enhancement approach achieved further reductions in response harmfulness.

\begin{table*}[t]
\centering
\resizebox{\textwidth}{!}{
    \renewcommand\arraystretch{1.4} 
    \setlength{\tabcolsep}{3.0mm}{} 
    \begin{tabular}{l|c|ccc}
    \hline
        \textbf{} & \textbf{Risk Content Identification} & \multicolumn{3}{c}{\textbf{Refusal to Answer}}\\ 
         & ACC & RR-1 & RR-2 & HR  \\ \hline
         
\textbf{Base Models} & 76.17\% & 60.65\% & 55.58\% & 5.58\% \\ \hline

\textbf{Distilled Models} & 74.00\% & 59.36\% & 50.65\% & 3.13\%  \\ \hline

    \textbf{Safety-enhanced Distilled Models} & \textbf{83.12\%} & \textbf{66.92\%} & \textbf{65.88\%} & \textbf{1.55\%}  \\ \hline
    \end{tabular}
    }
\caption{Overall evaluation metrics of DeepSeek-R1 distilled model series, its corresponding base model series, and safety-enhanced model series across two tasks. Higher ACC, RR-1 and RR-2 are indicative of better performance, whereas lower HR is preferable. The optimal values under the current metric are highlighted bold.}
\label{tab:question-type}
\end{table*}

\begin{table*}[t]
\centering
\resizebox{\textwidth}{!}{
    \renewcommand\arraystretch{1.4} 
    \setlength{\tabcolsep}{3.0mm}{} 
    \begin{tabular}{l|c|ccc}
    \hline
        \textbf{} & \textbf{Risk Content Identification} & \multicolumn{3}{c}{\textbf{Refusal to Answer}}\\ 
         & ACC & RR-1 & RR-2 & HR  \\ \hline

\textbf{R1 Models} & 73.63\% & 60.54\% &53.01\% & 2.78\%  \\ \hline

\textbf{Safety-enhanced R1 Models} & \textbf{84.35\%} & \textbf{68.93\%} & \textbf{67.73\%} & \textbf{1.33\%}  \\ \hline
    
    \end{tabular}
    }
\caption{Overall evaluation metrics of DeepSeek-R1 model series, and safety-enhanced model series across two tasks. Higher ACC, RR-1 and RR-2 are indicative of better performance, whereas lower HR is preferable. The optimal values under the current metric are highlighted bold.}
\label{tab:question-type}
\end{table*}

Table 4 presents a comparative analysis of the overall safety capabilities across all seven DeepSeek-R1 models before and after safety enhancement. In the risk content identification task, the safety-enhanced models demonstrated a 10.72\% improvement in accuracy compared to their non-enhanced counterparts. For the refusal-to-answer task, the enhanced models showed significant gains of 18.39\% in RR-1 and 14.72\% in RR-2 metrics, along with a further 1.45\% reduction in HR, achieving comprehensive safety improvement. Furthermore, as evidenced in Table 3, the safety-enhanced models not only outperformed their non-enhanced versions but also surpassed the base models (which originally exhibited slightly stronger safety capabilities) in several aspects: the accuracy in risk content identification improved by 6.95\% over base models; the RR-1 and RR-2 metrics increased by 6.27\% and 10.30\% respectively in refusal-to-answer tasks; and the HR metric decreased by 4.03\%, representing a complete outperformance of the base models across all evaluated dimensions.

In summary, the safety-enhanced models demonstrate significantly superior safety capabilities compared to both the non-enhanced models and the non-distilled base models. At the same time, the reasoning capabilities of the safety-enhanced models remain stable, with no significant decline observed, and even improvements in some benchmark tests. These results robustly validate the effectiveness of the safety enhancement strategy.

\section{Conclusion}

In response to widespread societal concerns regarding the safety issues of DeepSeek-R1, this paper conducts a systematic safety evaluation of the entire R1 series of distilled models and quantitatively analyzes the negative impact of the distillation process on the safety capabilities of the non-distilled base models. Building on this analysis, the paper further implements targeted safety enhancement measures for all DeepSeek-R1 models to rapidly improve their safety performance. Comprehensive evaluation results demonstrate that the safety-enhanced DeepSeek-R1 models achieve a significant safety improvement, even surpassing the base models with stronger safety capabilities, while exhibiting no notable negative impact on reasoning performance.

%
%
%
\bibliographystyle{splncs04}
\bibliography{main}



\end{document}